\documentclass[10pt,twocolumn,letterpaper]{article}

\usepackage{iccv}
\usepackage{times}
\usepackage{epsfig}
\usepackage{graphicx}
\usepackage{amsmath}
\usepackage{amssymb}

\usepackage{cite}
\usepackage{amsmath,amssymb,amsfonts}
\usepackage{algorithmic}
\usepackage{graphicx}
\usepackage{textcomp}
\usepackage{xcolor}

\usepackage{graphics} %
\usepackage{multicol,multirow}
\usepackage{subfigure}
\usepackage{multirow}
\usepackage{setspace}

\newcommand{\Rong}[1]{\textcolor{black}{#1}}
\newcommand{\paragraphcond}[1]{\vspace{0.3em}\noindent\textbf{#1}}

\graphicspath{{./figs/}}
\DeclareGraphicsExtensions{.pdf,.jpg,.png}

\newcommand{\figref}[1]{Fig.~\ref{#1}}
\newcommand{\tabref}[1]{Table~\ref{#1}}

\usepackage[pagebackref=true,breaklinks=true,letterpaper=true,colorlinks,bookmarks=false]{hyperref}

\iccvfinalcopy %

\def\iccvPaperID{15} %

\ificcvfinal\fi

\begin{document}

\title{Semantic RGB-D Image Synthesis
}

\author{Shijie Li\\
University of Bonn\\
Bonn, Germany\\
{\tt\small lishijie@iai.uni-bonn.de}
\and
Rong Li\\
HKUST (Guangzhou)\\
Guangzhou, China\\
{\tt\small rongli@hkust-gz.edu.cn}
\and
Juergen Gall\\
University of Bonn\\
Bonn, Germany\\
{\tt\small gall@iai.uni-bonn.de}
}

\maketitle
\ificcvfinal\thispagestyle{empty}\fi

\begin{abstract}
Collecting diverse sets of training images for RGB-D semantic image segmentation is not always possible. In particular, when robots need to operate in privacy-sensitive areas like homes, the collection is often limited to a small set of locations. As a consequence, the annotated images lack diversity in appearance and approaches for RGB-D semantic image segmentation tend to overfit the training data. In this paper, we thus introduce semantic RGB-D image synthesis to address this problem. It requires synthesising a realistic-looking RGB-D image for a given semantic label map. Current approaches, however, are uni-modal and cannot cope with multi-modal data. Indeed, we show that extending uni-modal approaches to multi-modal data does not perform well. In this paper, we therefore propose a generator for multi-modal data that separates modal-independent information of the semantic layout from the modal-dependent information that is needed to generate an RGB and a depth image, respectively. Furthermore, we propose a discriminator that ensures semantic consistency between the label maps and the generated images and perceptual similarity between the real and generated images. Our comprehensive experiments demonstrate that the proposed method outperforms previous uni-modal methods by a large margin and that the accuracy of an approach for RGB-D semantic segmentation can be significantly improved by mixing real and generated images during training.   
\end{abstract}

\section{Introduction}

RGB-D semantic segmentation is essential for mobile agents as it enables gaining a precise perception of the environment. While there are fast methods like ESANet~\cite{seichter2021efficient} that are suitable for robotics applications, the collection of annotated RGB-D training data remains a bottleneck. This is in particular an issue for robots in homes since the data collection is restricted due to privacy concerns, but it is also an issue for UAVs or delivery robots that need to enter private property.

An interesting direction to increase the diversity of training data is thus to generate training data from label maps as shown in Fig.~\ref{fig:qvis_stoa}. This task is also known as semantic image synthesis~\cite{park2019semantic,schonfeld_sushko_iclr2021, SushkoSZGSK22} where a semantic label map is provided and the aim is to generate realistic images that are consistent with the label map. While this also requires training data, it allows the generation of each annotated training image variant where the appearance of the present objects, walls, and floor differs from the original image as shown in Fig.~\ref{fig:qvis_stoa}. While previous works focused on uni-modal semantic image synthesis, i.e., only generating RGB images, we propose a multi-modal approach that generates realistic-looking RGB-D images from semantic label maps. We thus call the task semantic RGB-D image synthesis.  
In contrast to previous works, we not only evaluate the quality of the generated images but also demonstrate that we can significantly improve the accuracy of an RGB-D semantic segmentation approach~\cite{seichter2021efficient} by mixing real and generated images during training as shown in Fig.~\ref{fig:combined}.

\begin{figure}[t]
    \centering
    \subfigure[Real]{
    \begin{minipage}{0.32\linewidth}
    \centering
    \includegraphics[width=\linewidth,height=1.8cm]{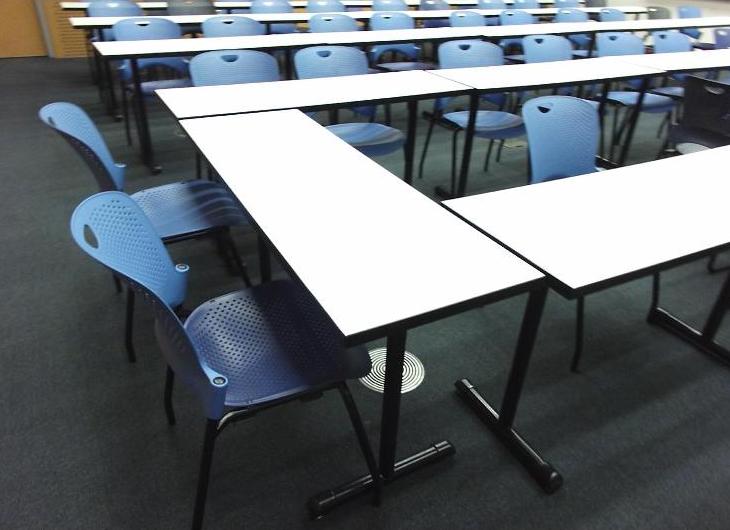} \\ 
    \includegraphics[width=\linewidth,height=1.8cm]{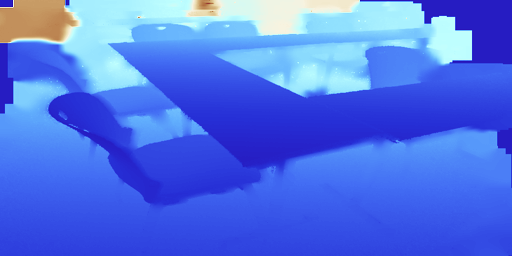} \\
    \vspace{1mm}
    \end{minipage}} \hspace{-3mm}
    \subfigure[RGB-D OASIS]{
    \begin{minipage}{0.32\linewidth}
    \centering
    \includegraphics[width=\linewidth,height=1.8cm]{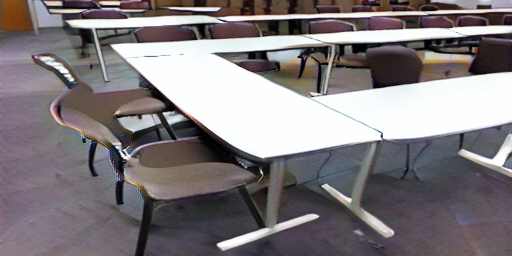} \\ 
    \includegraphics[width=\linewidth,height=1.8cm]{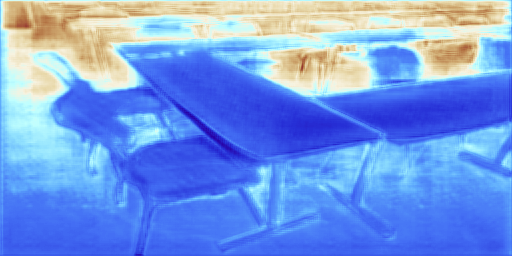} \\ 
    \vspace{1mm}
    \end{minipage}} \hspace{-3mm}
    \subfigure[SCMIS \Rong{(Ours)}]{
    \begin{minipage}{0.32\linewidth}
    \centering
    \includegraphics[width=\linewidth,height=1.8cm]{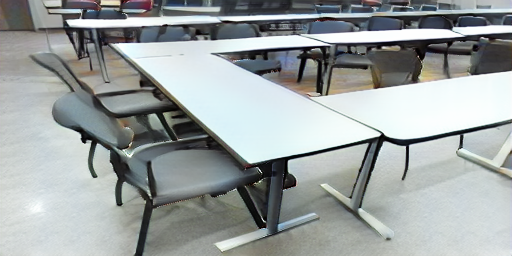} \\ 
    \includegraphics[width=\linewidth,height=1.8cm]{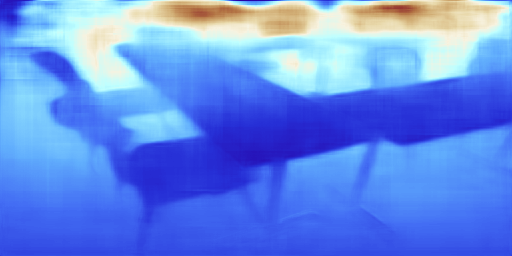} \\ 
    \vspace{1mm}
    \end{minipage}}
    \caption{Comparison of images that are generated for a given label mask. The left column shows the real RGB-D image that corresponds to the label mask. The middle column shows RGB-D images generated by OASIS extended to RGB-D. The proposed approach (right column) generates more realistic RGB-D images than OASIS. While the depth maps of OASIS are inaccurate and noisy and the edges of the tables are not straight, our approach generates images where the RGB and depth image are consistent.
    }
    \vspace{-6mm}
    \label{fig:examples}
\end{figure}

Although approaches for uni-modal semantic image synthesis like OASIS~\cite{schonfeld_sushko_iclr2021} can also be applied to RGB-D images by treating RGB and depth as a single modality, they do not perform well as shown in \figref{fig:examples}. We therefore propose an approach that explicitly considers RGB and depth as two different modalities. However, since appearance and geometry are highly correlated, we aim to disentangle the modal-independent information that is encoded in the semantic label map and the modal-dependent information that is needed to generate images for both modalities, namely depth and color. 

\begin{figure*}[!t]
    \centering
\resizebox{0.9\linewidth}{!}{
    \includegraphics{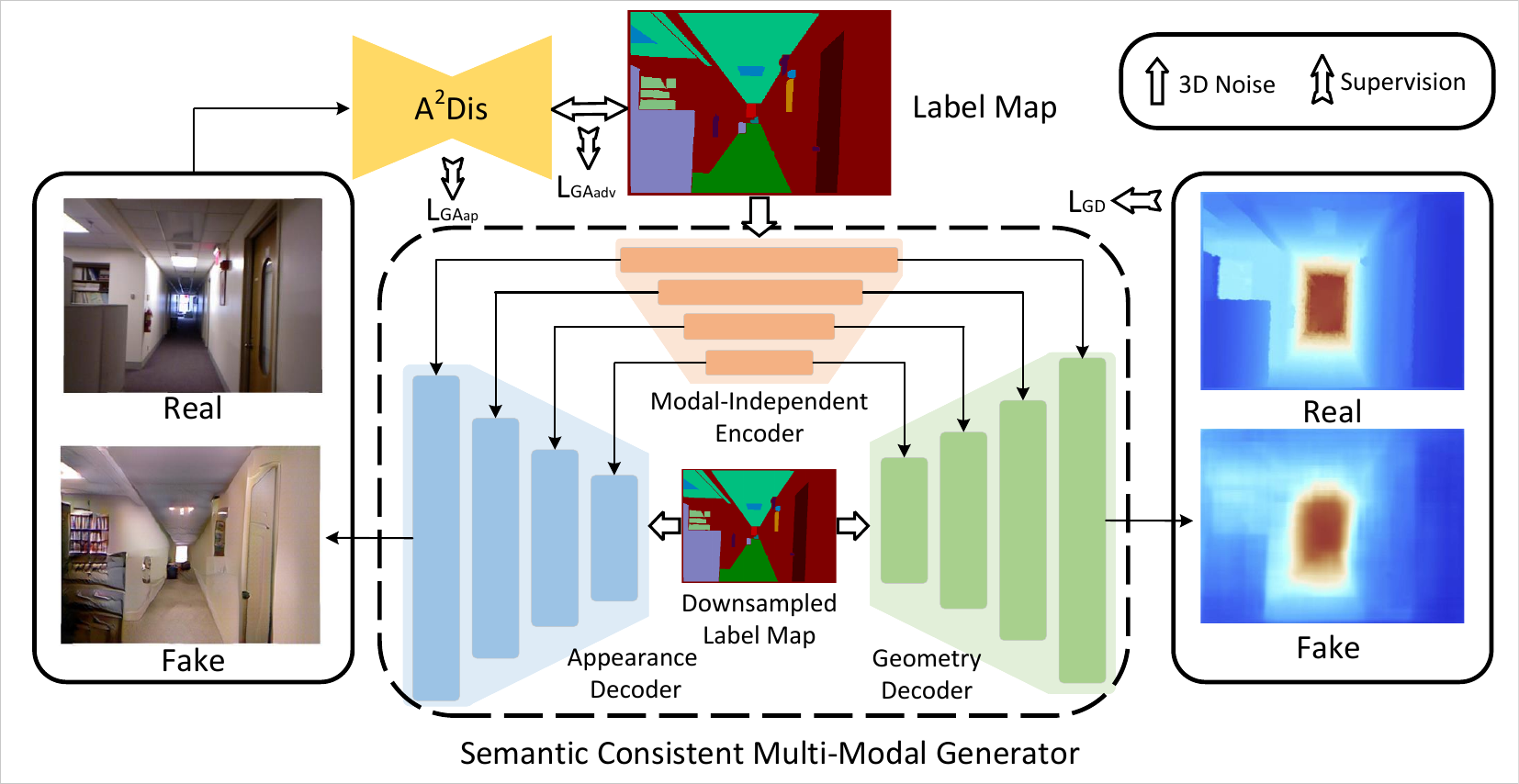}
    }
    \caption{Overview of the proposed \textbf{S}emantic \textbf{C}onsistent \textbf{M}ulti-Modal \textbf{I}mage \textbf{S}ynthesis (SCMIS) approach. The generator takes a label map as input and generates an RGB image (blue) and depth image (green) that are semantically consistent with the input label map. For training, we use for each modality a different loss term. While $\mathcal{L}_{GD}$ measures the quality of the generated depth map, the proposed Adaptive Alignment Discriminator (A$^2$Dis) measures the quality of the generated RGB image and consistency with the input label map.} 
    \label{fig:arch}
\end{figure*}

To this end, the generator uses an encoder that is modal-independent and two modal-dependent decoders for both modalities as illustrated in \figref{fig:arch}. The decoding is done gradually from the down-sampled label map with spatially-adaptive normalization, which is conditioned on the modal-independent information for each scale. To train the model, we ensure that the generated depth maps correspond to the real depth maps. Furthermore, we propose a discriminator that ensures semantic consistency between the generated image and the semantic label map. We call the approach thus \textbf{S}emantic \textbf{C}onsistent \textbf{M}ulti-Modal \textbf{I}mage \textbf{S}ynthesis (SCMIS).

Since we also aim that the RGB images to look as realistic as possible, we enforce that the features of generated and real images are similar. While this can be theoretically achieved by the perceptual loss \cite{johnson2016perceptual}, we show that the perceptual loss impedes learning and has a negative impact on the results. This is due to the commonly used VGG features that do not necessarily separate the semantic classes. We, therefore, integrate the perceptual loss into the discriminator such that the VGG features are adapted and separate the classes better.   

We evaluate our approach for multi-modal semantic image synthesis on standard RGB-D datasets where it outperforms uni-modal approaches by a large margin. 
Furthermore, we show that the generation of depth also improves the quality of the generated RGB images. The contributions of this work can be summarized as:
1) We propose a novel semantic consistent multi-modal generator that generates realistic RGB-D images by disentangling the modal-independent and modal-dependent information.
2) We propose an adaptive alignment discriminator which performs well for uni-modal (RGB) and multi-modal (RGB-D) semantic image synthesis.
3) We conduct comprehensive experiments and the results demonstrate that the proposed method outperforms previous methods by a large margin.
4) We demonstrate that the accuracy of an RGB-D semantic segmentation approach can be improved by mixing the real training images with the generated images.

\section{Related Work}

\textbf{Generative Adversarial Networks.} GAN\cite{goodfellow2014generative}
are commonly used for semantic image synthesis and many improvements have been proposed over the years.    
One main issue is the instability in adversarial training.
\cite{radford2015unsupervised} improved the training stability by designing a robust architecture and providing some design guidelines.
Different from \cite{radford2015unsupervised}, some methods \cite{arjovsky2017wasserstein,salimans2016improved,mao2017least} made the training more stable by using some tricks like adding regularizers or proposing better loss functions. In addition, one can easily improve the efficiency of networks by existing model compression techniques~\cite{xu2023resilient, xu2022ida}.

\textbf{Semantic Image Synthesis.} 
Pix2Pix \cite{isola2017image} was one of the first approaches for semantic image synthesis and used a general image-to-image translation framework with a conditional GAN architecture \cite{mirza2014conditional}. Since Pix2Pix \cite{isola2017image} can only handle images of the small resolution, Pix2PixHD \cite{wang2018high} proposed a multi-resolution approach to generate high-resolution images.
SPADE \cite{park2019semantic} removed the common normalization layers like  instance normalization \cite{ulyanov2016instance} and instead modulates the image content spatially by a learned affine transformation.
Inspired by SPADE \cite{park2019semantic}, many semantic image synthesis methods have been proposed \cite{tang2020dual,tang2020local,jiang2020tsit,schonfeld_sushko_iclr2021}.
Among these methods, OASIS \cite{schonfeld_sushko_iclr2021} showed that a segmentation-based discriminator provides much more precise supervision and improves the results substantially.

\textbf{Geometric Image Synthesis.}
Recently, some works explore 3D data generation given 2D supervision.
Most of them focus on specific shapes like faces~\cite{piao2021inverting,gecer2021fast,gecer2019ganfit}.
Some works go further \cite{pan20202d,hu2021self} and reconstruct some simple 3D models from a single image. 
Apart from reconstructing 3D models, some methods generate geometric-consistent images.
For instance, \cite{zhu2018generative} generates a bird view from a frontal view in the context of a driving car.
\cite{wu2019transgaga,fu2019geometry} utilize geometric information to improve the quality of generated images. While these works generate geometry or utilize geometry for generating images, none of them generates multi-modal data. We will also show that uni-modal approaches for semantic image synthesis do not perform well for semantic RGB-D image synthesis.

\begin{figure*}[t]
\centering
\subfigure[VGG-19 pretrained on ImageNet \cite{krizhevsky2012imagenet}]{
\begin{minipage}[t]{0.49\linewidth}
\centering
    \includegraphics[width=0.49\linewidth]{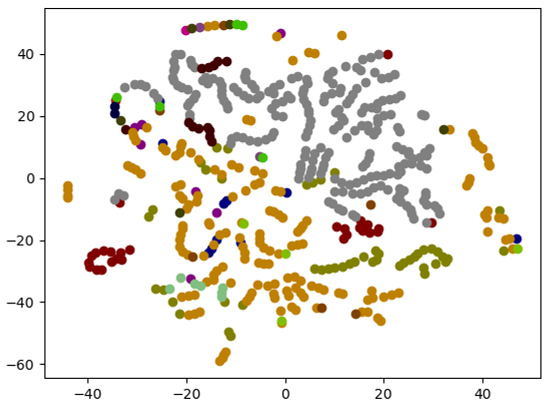} 
    \includegraphics[width=0.49\linewidth]{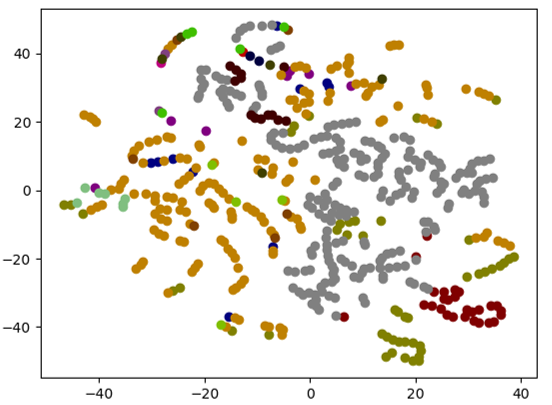} 
\end{minipage}
}  \hspace*{-3mm}
\subfigure[Adaptative VGG-19]{
\begin{minipage}[t]{0.49\linewidth}
\centering
    \includegraphics[width=0.49\linewidth]{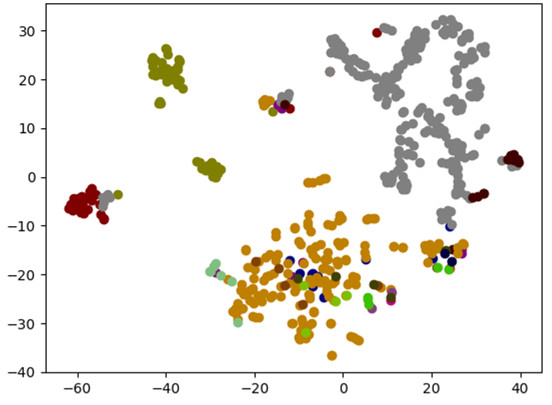} 
    \includegraphics[width=0.49\linewidth]{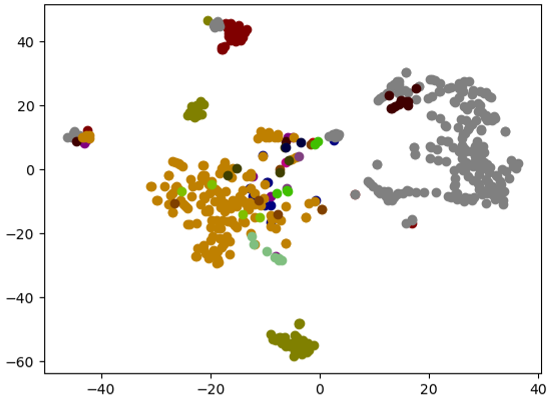} 
\end{minipage}
}  \hspace*{-3mm}
\caption{(a) Visualization of the features that are used for the perceptual loss, i.e., VGG-19 pre-trained on ImageNet \cite{krizhevsky2012imagenet}. The features are projected by t-SNE \cite{van2008visualizing}. The left plot shows the features for real images and the right plot for generated images. The colors correspond to different semantic classes. (b) Visualization of the features that are learned for adaptive perceptual loss. 
}
\label{fig:feat_vis}
    \vspace{-5mm}
\end{figure*}

\section{Semantic Consistent Multi-Modal Image Synthesis}

While previous works for semantic image synthesis generate only one modality, namely an RGB image, for a given label map, we address multi-modal semantic image synthesis. Instead of generating only an RGB image from a given label map, our approach generates an RGB-D image where color and depth are treated as two modalities as shown in \figref{fig:arch}. An important aspect of the architecture is that modal-independent information of the semantic layout and  modal-dependent information that is needed to generate an RGB and a depth image are separated.

\subsection{Semantic Consistent Multi-Modal Generator}\label{sec_gen}

Previous methods for semantic image synthesis considered only one modality and naively extending these methods to multi-modal data does not result in good results as shown in \figref{fig:examples}. Indeed, we will demonstrate in the experiments that generating RGB-D images from semantic label maps in the same way as generating RGB images is even slightly worse than generating both modalities RGB and depth independently. However, depth and color are highly correlated and should not be treated independently. We therefore address the research question of how RGB-D images can be better generated from semantic label maps.

Our proposed multi-modal generator shown in \figref{fig:arch} takes into account that both modalities share the same scene properties like layout or object locations, which are independent of the modality.
The modal-independent encoder (red) thus encodes the common properties that are shared by both modalities. The decoders, however, are modal-dependent and we use one appearance decoder (blue) and one geometry decoder (green), which generate the RGB image and the depth image, respectively. We now describe the encoder and both decoders more in detail.   

As shown in \figref{fig:arch}, the modal-independent encoder takes as input a channel-wise concatenation of a semantic label map and a 3D noise tensor. The noise tensor allows to generate various plausible images from the same label map. The encoder consists of blocks with a convolution layer, a batch normalization layer, and a ReLU. The resolution decreases after each block.

The modal-dependent decoders have the same architecture and generate from a downsampled version of the semantic map concatenated with the 3D noise tensor the RGB image or depth map, respectively, at the resolution of the original input label map. Each decoder consists of ResNet blocks \cite{he2016deep} with spatially-adaptive normalization (SPADE) \cite{park2019semantic}:
\begin{equation}
    \gamma^i_{x,y,c}(\mathbf{e}^i)\frac{h^i_{x,y,c,n}-\mu^i_c}{\sigma^i_c} + \beta^i_{x,y,c}(\mathbf{e}^i)
    \label{eq:spade}
\end{equation}
where  $n$ is the batch size, $c$ is the channel, $(x, y)$ is a pixel at layer $i$. $h^i_{x,y,c,n}$ is the input to the normalization layer, and 
$\mu^i_c$ and $\sigma^i_c$ are the mean and standard deviation of $h^i_{x,y,c,n}$ for channel $c$, i.e., computed over $n$, $x$, and $y$. The spatially-adaptive normalization can be conditioned on a tensor $\mathbf{e}$.

We use the tensor $\mathbf{e}$ to inject the modal-independent information at each scale for the decoders. $\mathbf{e}^i$ is thus the tensor of the encoder at layer $i$. In this way, both decoders share the same semantic and modal-independent information and gradually generate from the downsampled input map the two different modalities.

\subsection{Adaptive Alignment Discriminator}\label{sec_disc}

Before we describe the full loss to train the generator, we first discuss the second contribution, which is denoted by Adaptive Alignment Discriminator (A$^2$Dis) in \figref{fig:arch}. The discriminator serves two purposes. It aims to ensure the semantic consistency of the generated image with the label map and aims to ensure that the generated images look realistic. 

To this end, we use a discriminator that predicts per-pixel probabilities of $N+1$ classes which correspond to $N$ semantic classes and one additional `fake' class as in  \cite{schonfeld_sushko_iclr2021}. For the discriminator, we will evaluate different types in our experiments as shown in \figref{fig:dis}. The network is trained by using the ground-truth label map for a real image and labelling all pixels as `fake' for a generated image. To deal with imbalanced classes, a weighted $N+1$-class cross-entropy loss is used:
\begin{equation}
\begin{split}
        \mathcal{L}_{GA_{adv}} = -\mathbf{E}_{(\mathbf{x}, \mathbf{l})}\left[\sum_{c=1}^N\alpha_c \sum_{x,y}^{H\times W}\mathbf{l}_{x,y,c} \log D(\mathbf{x})_{x,y,c}\right] 
        \\ - \mathbf{E}_{(\mathbf{z},\mathbf{l})}\left[\sum_{x,y}^{H\times W}\log D(G(\mathbf{z},\mathbf{l}))_{x,y,c=N+1}\right]
\end{split}
    \label{eq:l_d_seg}
\end{equation}
where $\mathbf{l}$ is the semantic label map,  $\mathbf{x}$ is a real image, and $(\mathbf{z}, \mathbf{l})$ is the concatenation of the 3D noise tensor and the semantic label map. The weight $\alpha_c$ is computed as the inverse of the per-pixel class frequency to give rare classes higher weights:
\begin{equation}
    \alpha_c = E_{\mathbf{l}}\left[\frac{H\times W}{\sum_{x,y}^{H\times W} \mathbf{l}_{x,y,c}}\right].
\end{equation}
By this design, the semantic consistency among generated images are enforced.

Although this discriminator can train a good generator, it does not ensure that the generator produces images that follow the real data distribution as no constraint between generated images and real images exists.
To solve this issue, we can complement our discriminator with the perceptual loss \cite{johnson2016perceptual}, which is widely used for training generative networks \cite{park2019semantic}. But this will introduce two issues.
First, an additional pre-trained VGG-19 model is required which is inefficient.
Second, the used VGG-19 model is pre-trained on another dataset and might not be suitable for our task. This domain gap leads to semantic inseparable features as can be seen in \figref{fig:feat_vis} (a). As consequence, the perceptual loss works against the loss \eqref{eq:l_d_seg} since it pushes the generated features towards a feature space where the classes are not separable and difficult to classify.

The proposed adaptive alignment discriminator solves these issues by unifying the semantic consistency among generated images and the alignment between real and generated images in the same framework. It is designed as a segmentation-based discriminator with a learnable VGG-19 as the backbone. In this way, the learned features fit better to the task and separate the semantic classes better as shown in \figref{fig:feat_vis} (b). To maintain feature alignment between real and generated images, the perceptual loss is applied to the learned features, and thus named adaptive perceptual loss:
\begin{equation}\label{eq:perc}
    \mathcal{L}_{GA_{ap}} = \frac{1}{N}\sum_{i=1}^{N} ||r_i - f_i||_1
\end{equation}
where $i$ denotes the layer number, $N$ is the total number of layers, and $r_i$ and $f_i$ denote the features from the real and generated image, respectively. 

As mentioned at the beginning of this section, we investigated three architectures for the discriminator with the adaptive VGG-19 backbone (AVGG). As shown in \figref{fig:dis}, it includes a simple architecture with upsampling, pyramid pooling \cite{zhao2017pyramid}, and a U-Net architecture \cite{ronneberger2015u}. As we will show in the experiments, the architecture with pyramid pooling performs best. Hence, the total loss for the appearance decoder is given by:
\begin{equation}
    \mathcal{L}_{GA} = \mathcal{L}_{GA_{adv}} + \mathcal{L}_{GA_{ap}}.
\end{equation}

To recover the geometric structure, we apply the L1 loss on the output of the geometry decoder:
\begin{equation}
    \mathcal{L}_{GD} = \frac{1}{N}\sum_{i=1}^N||d_i - \hat{d}_i||_1
\end{equation}
where $N$ is the number of pixels, $\hat{d}_i$ is the generated depth value, and $d_i$ is the ground truth depth value for pixel $i$.
In addition, 
we also include the LabelMix regularization \cite{schonfeld_sushko_iclr2021}:
\begin{equation}
\begin{split}
    \mathcal{L}_{LM} = &||D_{logits}(LabelMix(\mathbf{x}, \hat{\mathbf{x}}, M)) \\
    & - LabelMix(D_{logits}(\mathbf{x}), D_{logits}(\hat{\mathbf{x}}), M)||_2^2
\end{split}
\end{equation}
where $D_{logits}$ are the logits before the last softmax and $M$ is a binary mask for mixing real and generated images $(\mathbf{x}, \hat{\mathbf{x}})$ by 
\begin{equation}
    LabelMix(\mathbf{x}; \hat{\mathbf{x}};M) = M \odot \mathbf{x} + (1-M) \odot \hat{\mathbf{x}}.
\end{equation}
The entire loss is thus given by 
\begin{equation}
     \mathcal{L} = \mathcal{L}_{GA} + \mathcal{L}_{GD} + \mathcal{L}_{LM}.
\end{equation}

\begin{figure*}[t]
\centering
\subfigure[Label]{
\begin{minipage}[t]{0.18\linewidth}
\centering
\includegraphics[width=\linewidth,height=1.8cm]{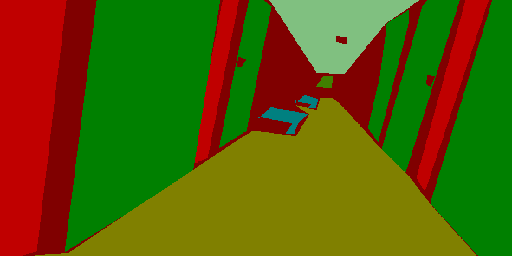} \\
\includegraphics[width=\linewidth,height=1.8cm]{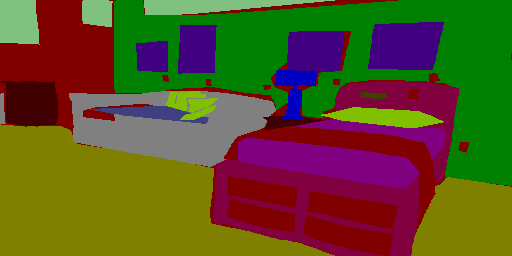} \\
\includegraphics[width=\linewidth,height=1.8cm]{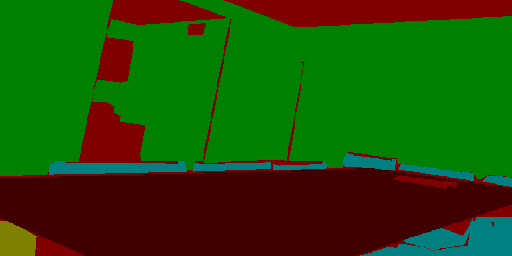} \\
\vspace{1mm}
\end{minipage}
}  
\hspace*{-3mm}
\subfigure[Real]{
\begin{minipage}[t]{0.18\linewidth}
\centering
\includegraphics[width=\linewidth,height=1.8cm]{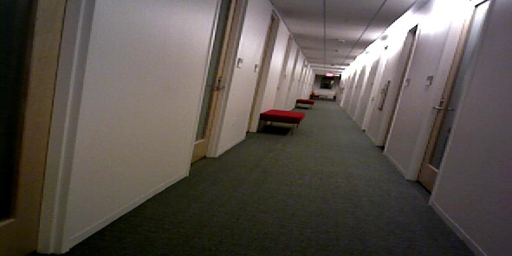} \\
\includegraphics[width=\linewidth,height=1.8cm]{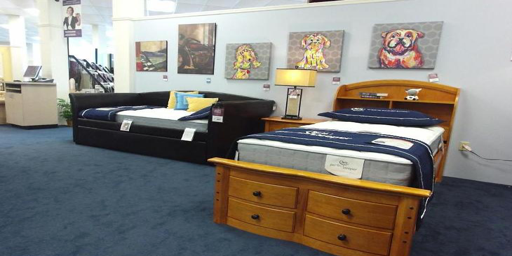} \\
\includegraphics[width=\linewidth,height=1.8cm]{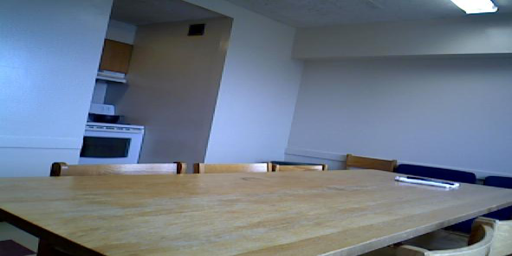} \\
\vspace{1mm}
\end{minipage}
} 
\hspace*{-3mm}
\subfigure[DAGAN~\cite{tang2020dual}]{
\begin{minipage}[t]{0.18\linewidth}
\centering
\includegraphics[width=\linewidth,height=1.8cm]{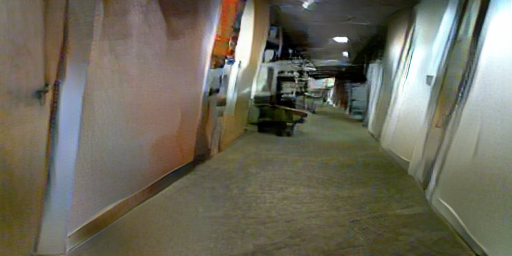} \\
\includegraphics[width=\linewidth,height=1.8cm]{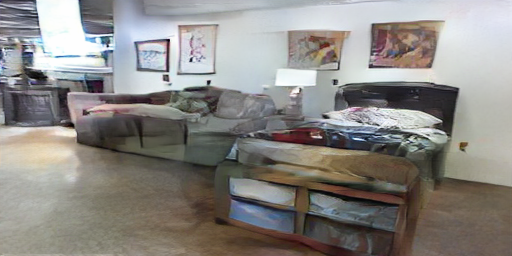} \\
\includegraphics[width=\linewidth,height=1.8cm]{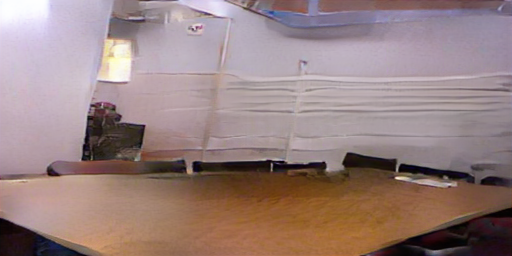} \\
\vspace{1mm}
\end{minipage}
} 
\hspace*{-3mm}
\subfigure[OASIS~\cite{schonfeld_sushko_iclr2021}]{
\begin{minipage}[t]{0.18\linewidth}
\centering
\includegraphics[width=\linewidth,height=1.8cm]{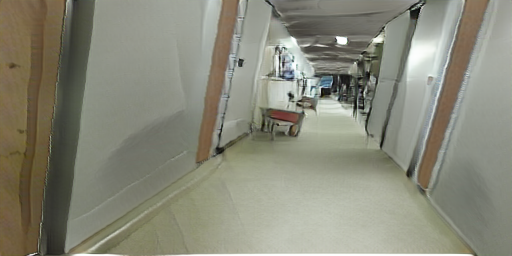} \\
\includegraphics[width=\linewidth,height=1.8cm]{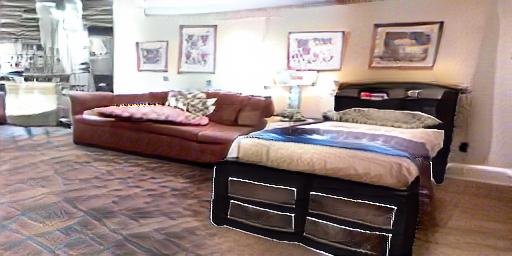} \\
\includegraphics[width=\linewidth,height=1.8cm]{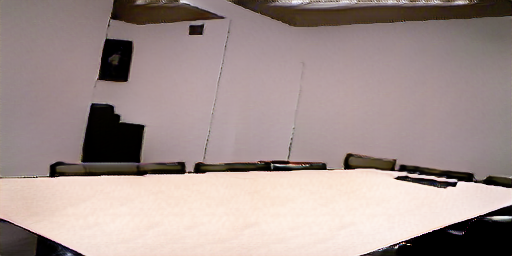} \\
\vspace{1mm}
\end{minipage}
} 
\hspace*{-3mm}
\subfigure[SCMIS (Ours)]{
\begin{minipage}[t]{0.18\linewidth}
\centering
\includegraphics[width=\linewidth,height=1.8cm]{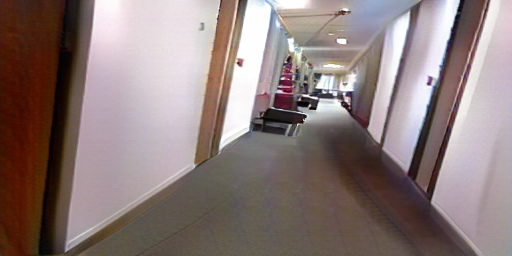} \\
\includegraphics[width=\linewidth,height=1.8cm]{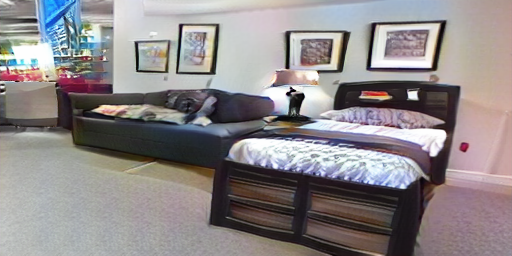} \\
\includegraphics[width=\linewidth,height=1.8cm]{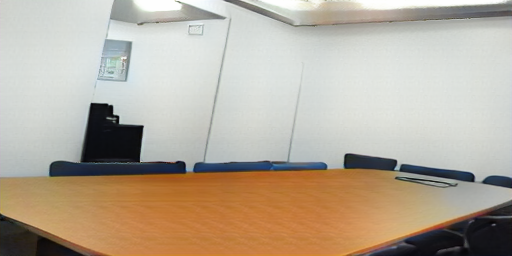} \\
\vspace{1mm}
\end{minipage}
}
\caption{Qualitative results of generated RGB images on the NYU dataset. The two most left columns show the input label map (a) and the corresponding real image (b). Compared to DAGAN and OASIS, our method can generate more realistic images. Besides better details and fewer artefacts, the illumination also looks more realistic due to the estimated geometry.      
}
\label{fig:qvis_stoa}
\vspace{-4mm}
\end{figure*}

\section{Experiment}

We evaluate our method on the two challenging datasets: SUN RGB-D \cite{song2015sun} and NYU \cite{silberman2012indoor}.
The NYU \cite{silberman2012indoor} dataset includes 1449 RGB-D images with 40 classes while the SUN RGB-D \cite{song2015sun} dataset contains 10355 RGB-D images with 37 classes.
We also evaluate our approach for uni-modal semantic image synthesis on the Cityscapes \cite{cordts2016cityscapes} dataset which includes 2975 training images and 500 validation images with 35 classes.
All experiments have been performed on a single TITAN RTX with a fixed random seed. The source code will be released upon acceptance. 

We will first evaluate the quality of the generated images in Section \ref{exp:synthesis} and compare our multi-modal approach to uni-modal approaches for semantic image synthesis. In Section \ref{exp:mix}, we will demonstrate that our approach can be used to augment the training data of an approach for RGB-D semantic segmentation by mixing real and generated images. Finally, we evaluate the impact of the loss terms and architecture design in Section~\ref{sec:ablation}.

\subsection{Semantic RGB-D Image Synthesis}\label{exp:synthesis}

\begin{figure}[t]
\centering
\subfigure[Real]{
\begin{minipage}[t]{0.31\linewidth}
\centering
    \includegraphics[width=\linewidth,height=1.8cm]{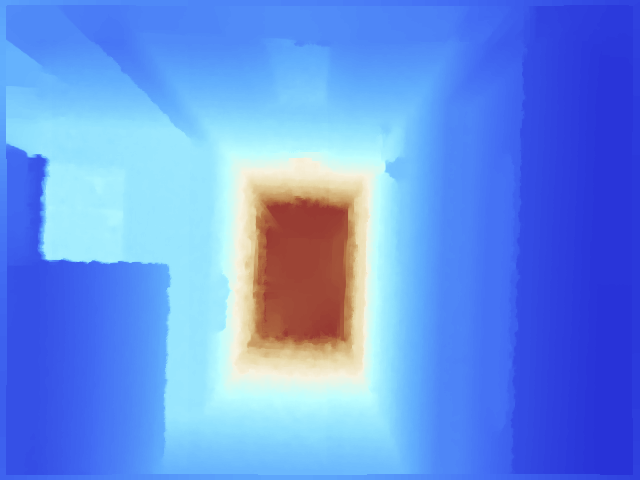} \\
    \includegraphics[width=\linewidth,height=1.8cm]{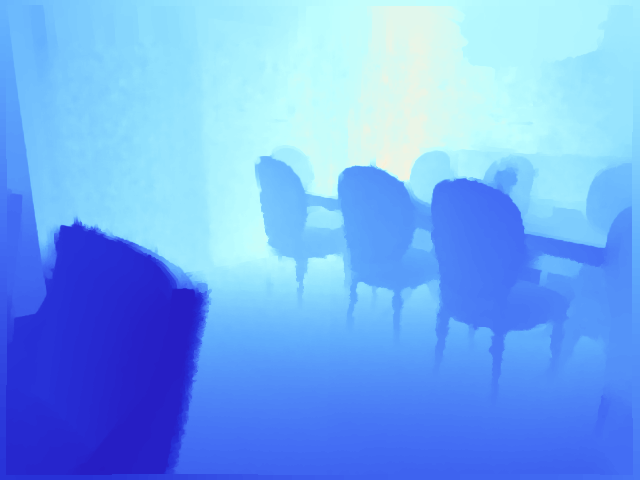} \\
    \includegraphics[width=\linewidth,height=1.8cm]{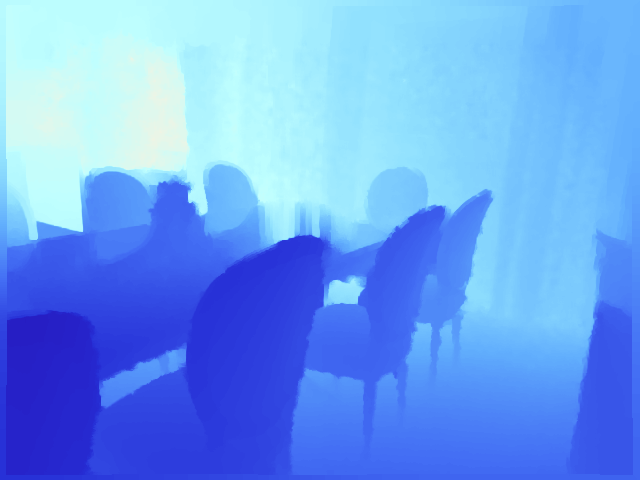} \\
\vspace{1mm}
\end{minipage}
}  \hspace*{-4mm}
\subfigure[Estimation~\cite{ranftl2021vision}]{
\begin{minipage}[t]{0.31\linewidth}
\centering
    \includegraphics[width=\linewidth,height=1.8cm]{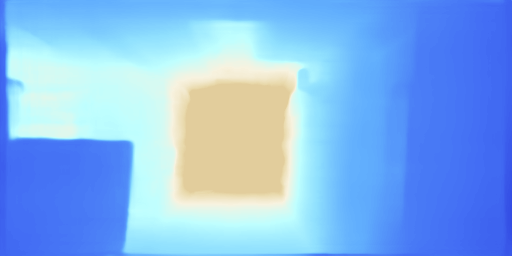} \\
    \includegraphics[width=\linewidth,height=1.8cm]{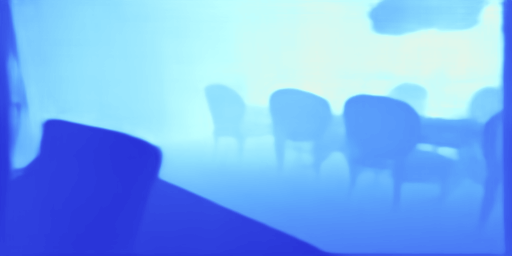} \\
    \includegraphics[width=\linewidth,height=1.8cm]{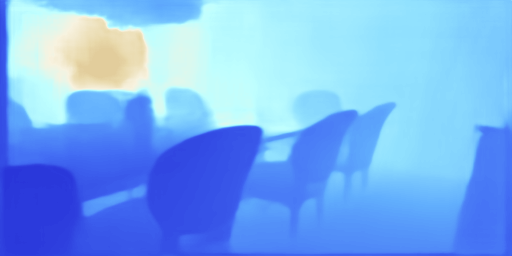} \\
\vspace{1mm}
\end{minipage}
} \hspace*{-4mm}
\subfigure[Generation (Ours)]{
\begin{minipage}[t]{0.31\linewidth}
\centering
    \includegraphics[width=\linewidth,height=1.8cm]{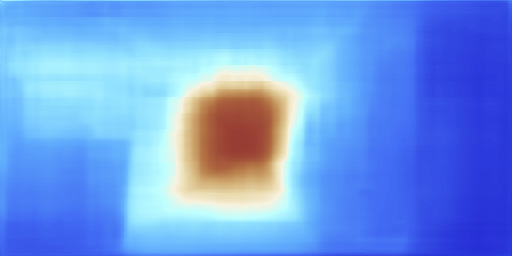} \\
    \includegraphics[width=\linewidth,height=1.8cm]{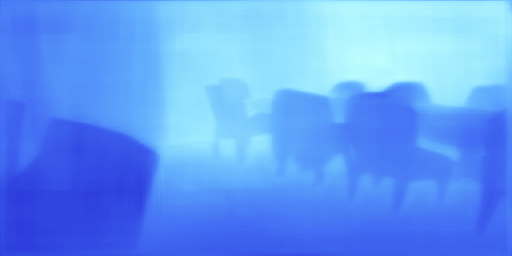} \\
    \includegraphics[width=\linewidth,height=1.8cm]{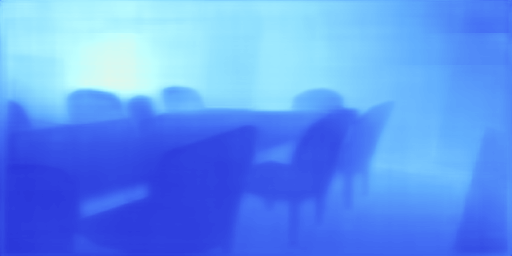} \\
\vspace{1mm}
\end{minipage}
}
\caption{Comparison of the real depth image (a) with the depth map that is estimated by \cite{ranftl2021vision} from the real color image (b) and the depth image that is generated by our method from the label map (c).   
}
\label{fig:qvis_depth}
\end{figure}

\begin{table}[t]
    \label{tab:comparasion_stoa_multi_modal}
    \centering
    \setlength{\tabcolsep}{1mm}
    \resizebox{0.8\linewidth}{!}{
    \begin{tabular}{l|cc|cc}
        \multirow{2}{*}{Methods}  &  \multicolumn{2}{c|}{NYU} & \multicolumn{2}{c}{SUNRGBD} \\
        \cline{2-5}
        & FID \textcolor{blue}{$\downarrow$}  & mIoU$_D$ \textcolor{red}{$\uparrow$} & FID \textcolor{blue}{$\downarrow$} & mIoU$_D$ \textcolor{red}{$\uparrow$}  \\
        \hline
        SPADE \cite{park2019semantic} & 122.1 & 47.2 & 246.6 & 27.8 \\
        TSIT \cite{jiang2020tsit} & 125.4 & 51.8 & 130.8 & 40.5\\
        DAGAN \cite{tang2020dual} & 99.2 & 51.4 & 57.4 & 51.2 \\
        OASIS \cite{schonfeld_sushko_iclr2021} & 77.5 & 58.3 & 26.4 & \textbf{62.2} \\
        RGB-D OASIS & 89.1 & 54.7 & 25.1 & 57.8 \\
        \hline
        SCMIS & \textbf{55.2} & \textbf{59.8} & \textbf{19.9} & 60.7\\
    \end{tabular}}
    \vspace{1mm}
    \caption{Comparison on multi-modal image synthesis.}
\end{table}

We use the common  Fr\'{e}chet Inception Distance (FID) \cite{heusel2017gans} to evaluate the quality of the generated images using the implementation from \cite{schonfeld_sushko_iclr2021}.
Lower FID means better image quality.
To measure the semantic alignment between the generated image and the semantic label map (semantic consistency), we use mean  Intersection-over-Union (mIoU).
To this end, we use the RGB-D semantic segmentation framework ESANet \cite{seichter2021efficient} with pre-trained weights. For the evaluation, the generated images with real-depth images are used as input.
For uni-modal semantic image synthesis, we use DRN \cite{yu2017dilated} to maintain consistency with previous works. As we use two networks for computing mean Intersection-over-Union, `mIoU$_D$' in the tables denotes that ESANet \cite{seichter2021efficient} for RGB-D images has been used and `mIoU' denotes that DRN \cite{yu2017dilated} for RGB images has been used.

We compare our method to other methods for uni-modal semantic image synthesis on two RGB-D datasets.
The results are shown in \tabref{tab:comparasion_stoa_multi_modal}.
We observe that our method produces more realistic images and achieves a lower FID score. Adapting uni-modal semantic image synthesis methods to semantic RGB-D image synthesis leads to worse performance (RGB-D OASIS). This means that these architectures cannot handle multi-modal information well. 
The qualitative results in \figref{fig:qvis_stoa} also show that our method generates more realistic images compared to previous methods. 

We also qualitatively compare the depth images that are generated by our method from the label maps with the depth images that are estimated by the mono depth estimation method \cite{ranftl2021vision} from the real color images in \figref{fig:qvis_depth}. The results show that our approach can infer reasonable depth maps only from semantic information. A quantitative comparison is given in \tabref{tab:semantic_consistency}.

\subsection{Generating Training Data for RGB-D Semantic Segmentation}\label{exp:mix}

\begin{figure}[t]
\centering
\subfigure[Label]{
\begin{minipage}[t]{0.31\linewidth}
\centering
    \includegraphics[width=\linewidth,height=1.8cm]{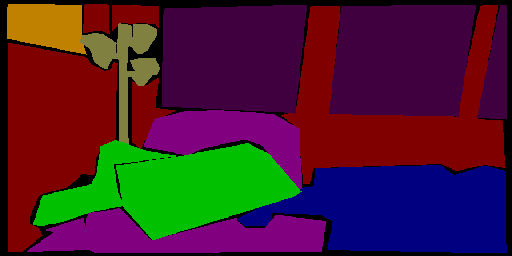} \\
    \includegraphics[width=\linewidth,height=1.8cm]{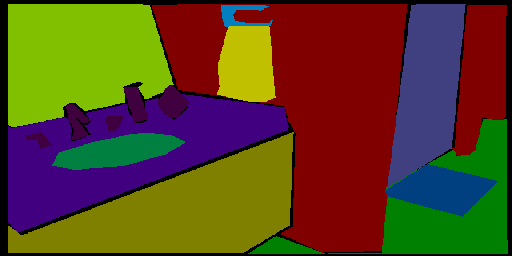} \\
    \includegraphics[width=\linewidth,height=1.8cm]{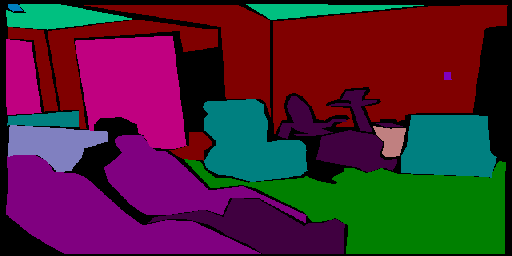} \\
\vspace{1mm}
\end{minipage}
}  \hspace*{-4mm}
\subfigure[Real]{
\begin{minipage}[t]{0.31\linewidth}
\centering
    \includegraphics[width=\linewidth,height=1.8cm]{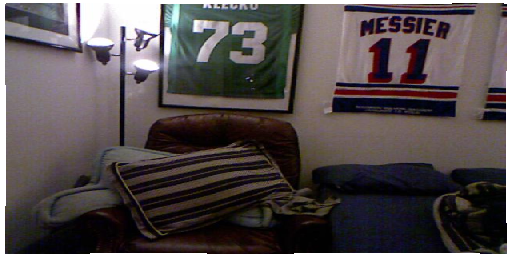} \\
    \includegraphics[width=\linewidth,height=1.8cm]{figs/combine/729_ori_img.png} \\
    \includegraphics[width=\linewidth,height=1.8cm]{figs/combine/428_ori_img.png} \\
\vspace{1mm}
\end{minipage}
} \hspace*{-4mm}
\subfigure[Mixed]{
\begin{minipage}[t]{0.31\linewidth}
\centering
    \includegraphics[width=\linewidth,height=1.8cm]{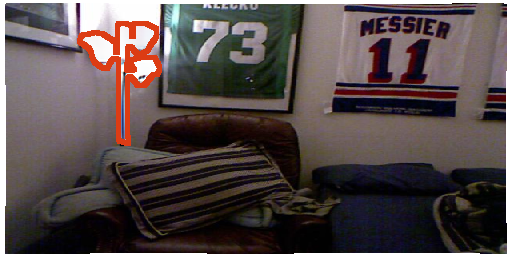} \\
    \includegraphics[width=\linewidth,height=1.8cm]{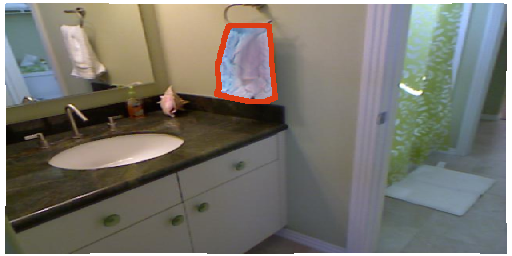} \\
    \includegraphics[width=\linewidth,height=1.8cm]{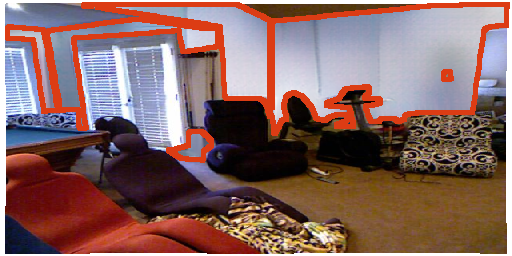} \\
\vspace{1mm}
\end{minipage}
}
\caption{Visualization of the semantic segmentation label (a) with the real color image (b) and the real and generated mixed color image (c), we highlight the generated pixels with red contour.}
\label{fig:combined}
\end{figure}

In \tabref{tab:aug_ratio}, we validate that an RGB-D semantic segmentation method can benefit from our generated data significantly. This is in particular relevant for robotics applications where annotated training data is limited. In this case, our approach can generate more variants for each annotated training image. Specifically, we train our method on the NYUv2~\cite{silberman2012indoor} training set and generate new RGB-D images for each semantic layout in the training set. We then mix the generated and original image by randomly picking some semantic classes and replacing the pixels for these classes in the original image with the pixels of the generated image as shown in \figref{fig:combined}. This generates the same dataset.
Finally, we train the RGB-D semantic segmentation method ESANet~\cite{seichter2021efficient} on the mixed images and evaluate the mIoU on the real images of the test set.

\tabref{tab:aug_ratio} shows the results for different percentages of randomly replaced classes. We furthermore evaluate if only depth, RGB values, or RGB-D values are replaced. The results show that all settings boost the performance significantly (4.89 - 6.16 mIoU), which demonstrates that the generated data increases the diversity of the dataset and improves the training of the baseline. The best results are achieved by replacing only RGB values, which is expected since appearance has more diversity than depth. Nevertheless, even replacing only depth values significantly improves the segmentation accuracy.

\begin{table}[t]

\label{tab:aug_ratio}
\centering
\setlength{\tabcolsep}{6mm}
\begin{tabular}{l|c|c}
Modality     & Ratio & mIoU \textcolor{red}{$\uparrow$} \\
\hline
-                      & 0.0              & 42.08          \\
\hline
\multirow{3}{*}{Depth} & 0.3            & 47.40 (+5.32)          \\
                       & 0.5            & 47.57 (+5.49)    \\
                       & 0.7            & 47.77 (+5.69)          \\
\hline
\multirow{3}{*}{RGB}   & 0.3            & 47.77 (+5.69)           \\
                       & 0.5            & 47.71 (+5.63)           \\
                       & 0.7            & \textbf{48.24 (+6.16)} \\
\hline
\multirow{3}{*}{RGB-D} & 0.3            & 46.97 (+4.89)          \\
                       & 0.5            & 47.85 (+5.77)         \\
                       & 0.7            & 47.84 (+5.76)        
\end{tabular}
\vspace{1mm}
\caption{Impact of mixing generated and real images on the NYUv2 dataset with image resolution 256x512. `-' denotes the baseline trained without generated images.
}
\end{table}

\begin{table}[t]
\label{tab:aug_methods}
\centering
\setlength{\tabcolsep}{2mm}
\begin{tabular}{c|ccc}
 Methods  & OASIS~\cite{schonfeld_sushko_iclr2021} & SCMIS (Only RGB) & SCMIS \\ \hline
 mIoU \textcolor{red}{$\uparrow$}   & 47.56 & 47.67 & \textbf{48.24}
\end{tabular}
\vspace{1mm}
\caption{Impact of different generation methods on the NYUv2 dataset. SCMIS (Only RGB) denotes SCMIS without geometry decoder.}

\end{table}

Since the best performance is achieved by only replacing RGB values, we also compare the results when we generate the data with a uni-modal generator. For this experiment, we use the setting where we replace 70\% of the classes. The results in \tabref{tab:aug_methods} show the benefit of multi-modal image generation. When we use the uni-modal method OASIS~\cite{schonfeld_sushko_iclr2021} or train SCMIS in a uni-modal setting, i.e., without the geometry decoder, the segmentation accuracy is lower. This is consistent with \tabref{tab:semantic_consistency}, which shows that the multi-modal approach generates more realistic images, i.e., lower FID score.

\subsection{Ablation Study}\label{sec:ablation}

\begin{figure*}[t]
\centering
\subfigure[Upsample]{
\begin{minipage}[t]{0.17\linewidth}
\centering
\includegraphics[width=\linewidth]{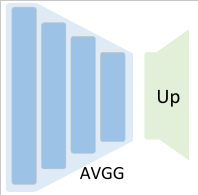} 
\end{minipage}
}  \hspace*{-2mm}
\subfigure[Pyramid Pooling]{
\begin{minipage}[t]{0.54\linewidth}
\centering
\includegraphics[width=\linewidth]{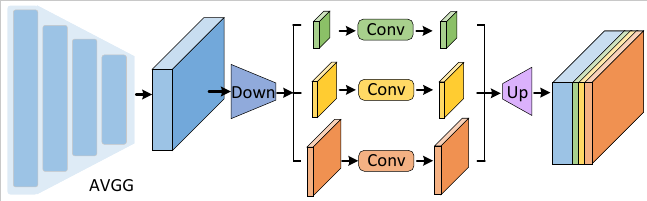} 
\end{minipage}
} \hspace*{-2mm}
\subfigure[U-Net]{
\begin{minipage}[t]{0.22\linewidth}
\centering
\includegraphics[width=\linewidth]{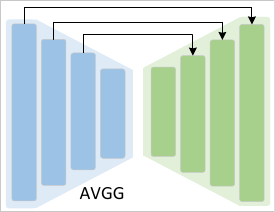} 
\end{minipage}
} 
\caption{Three architectures that are investigated for the discriminator. Each network uses an adaptive VGG-19 backbone (AVGG).
}
\label{fig:dis}
\end{figure*}

In this section, we analyse different design choices of the proposed approach. The experiments include RGB-D image synthesis (NYU dataset) and RGB image synthesis (Cityscapes dataset). For RGB image synthesis, we use the uni-modal generator OASIS \cite{schonfeld_sushko_iclr2021} in combination with our discriminator.

\paragraphcond{Architectures of Discriminator.}
In \tabref{tab:arch}, we show the impact of different architectures on the discriminator. The architectures Upsample, U-Net, and Pyramid Pooling (PP) are shown in \figref{fig:dis}. For this experiment, we do not use adaptive perceptual loss. Without multi-scale ability (Upsample), the quality of the generated images is relatively low (high FID). Using multi-scale architectures for the discriminator clearly improves the results (U-Net and PP).
Compared to U-Net, Pyramid Pooling improves the quality of the generated images by nearly 4-5 FID at the cost of a slightly lower semantic consistency (less than 1 mIoU). %
We thus use Pyramid Pooling (PP) for our discriminator. So far, the discriminator has only taken the RGB image as input, but not the depth map.
We, therefore, evaluate what happens if we provide the RGB-D images as input to the discriminator. We denote this setting by RGBD-PP. We can see that both image quality and semantic consistency drop drastically. This shows that considering depth and color as the same modality for the discriminator does not work well.    
Compared to the discriminator of OASIS, our discriminator is much more compact (nearly 4$\times$ smaller) and it results in more realistic images for both RGB-D image synthesis (FID is reduced more than 20) and RGB image synthesis (FID is reduced more than 4).

\begin{table}[t]
    \centering
    \label{tab:arch}
    \setlength{\tabcolsep}{2mm}
    \resizebox{0.9\linewidth}{!}{
    \begin{tabular}{l|cc|cc|c}
        \multirow{2}{*}{Methods}  &  \multicolumn{2}{c|}{NYU} & \multicolumn{2}{c|}{Cityscapes} & Size \\
        \cline{2-5}
        & FID \textcolor{blue}{$\downarrow$}  & mIoU$_D$ \textcolor{red}{$\uparrow$} & FID \textcolor{blue}{$\downarrow$} & mIoU \textcolor{red}{$\uparrow$} & (MB) \\
        \hline
        OASIS~\cite{schonfeld_sushko_iclr2021} & 77.5 & 58.3 & 47.7 & 69.3 & 22.2 \\
        \hline
        Upsample & 63.7 & 58.0 & 51.1 & 71.9 & \textbf{4.1} \\
        UNet & 59.6 & \textbf{59.6} & 48.2 & \textbf{73.1} & 5.1 \\
        PP & \textbf{55.7} & 59.2 & \textbf{43.2} & 72.2 & 6.5\\
        \hline
        RGB-D PP & 77.3 & 55.9 & N/A & N/A & 6.5\\
    \end{tabular}}
    \vspace{1mm}
    \caption{Impact of different discriminator architectures. PP denotes Pyramid Pooling and RGB-D PP is a discriminator with Pyramid Pooling but takes RGB-D images as input.}
\end{table}

\paragraphcond{Adaptive Perceptual Loss.}
So far, we have not used the adaptive perceptual loss \eqref{eq:perc}. We evaluate its impact for semantic RGB image synthesis in \tabref{tab:comparasion_stoa_single_modal}.
The adaptive perceptual loss ($\mathcal{L}_{A_{ap}}$) improves the image quality and reduces FID by 1.5, while mIoU remains the same. This is expected since the loss does not measure the semantic consistency with the label map, but the perceptual similarity to the real image. If we use the standard perceptual loss ($\mathcal{L}_{A_{p}}$ (IN)), the FID score even slightly increases. This is consistent with our observation that it pushes the generated features towards a feature space where the classes are not separable and difficult to classify as it is illustrated in \figref{fig:feat_vis}. This can be also addressed by training the VGG-19 features on the training set before training the generator ($\mathcal{L}_{A_{p}}$ (CS)). While this reduces the FID score, it does not perform as well as the adaptive perceptual loss. Furthermore, it requires training two separate networks, which is not very practical.

\begin{table}[t]
    \label{tab:comparasion_stoa_single_modal}
    \centering
    \setlength{\tabcolsep}{2mm}
    \resizebox{0.6\linewidth}{!}{
    \begin{tabular}{l|cc}
        \multirow{2}{*}{Methods}  &  \multicolumn{2}{c}{Cityscapes}  \\
        \cline{2-3}
        & FID \textcolor{blue}{$\downarrow$}  & mIoU \textcolor{red}{$\uparrow$} \\
        \hline
        PP & 43.2 & \textbf{72.2} \\
        PP + $\mathcal{L}_{A_{p}}$ (IN)& 43.4 & 72.1 \\
        PP + $\mathcal{L}_{A_{p}}$ (CS) & 42.1 & 72.1 \\
        PP + $\mathcal{L}_{A_{ap}}$ & \textbf{41.7} & 72.1 \\
    \end{tabular}}
    \vspace{1mm}
    \caption{Impact of perceptual loss. $\mathcal{L}_{A_{p}}$: perceptual loss with features pre-trained on ImageNet (IN) or Cityscapes (CS); $\mathcal{L}_{A_{ap}}$: adaptive perceptual loss.}
\end{table}

\begin{table}[t]

    \label{tab:semantic_consistency}
    \centering
    \setlength{\tabcolsep}{1mm}
    \resizebox{\linewidth}{!}{
    \begin{tabular}{l|c|c|c|ccc}
        \multirow{2}{*}{Methods}  & \multirow{2}{*}{Modality} & \multicolumn{5}{c}{NYU}  \\
        \cline{3-7}
        & & FID \textcolor{blue}{$\downarrow$} & mIoU$_D$ \textcolor{red}{$\uparrow$} & AbsRel \textcolor{blue}{$\downarrow$} & RMSE \textcolor{blue}{$\downarrow$} & SqRel \textcolor{blue}{$\downarrow$} \\
        \hline
        SCMIS & Depth & N/A & N/A & 0.172 & 0.613 & 0.142 \\
        SCMIS & RGB & 59.2 & N/A & N/A &  N/A &  N/A\\ %
        \hline
        RGB-D OASIS & RGB-D & 89.1 & 54.7 & 0.221 & 0.737 & 0.204 \\ %
        SCMIS & RGB-D & {55.7} & 59.2 & \textbf{0.166} & \textbf{0.599} & \textbf{0.133}  \\ %
        \hline
        4D Gen + $\mathcal{L}_{A_{ap}}$ & RGB-D & 65.8 & \textbf{60.0} & 0.197 & 0.662 & 0.170 \\
        SCMIS + $\mathcal{L}_{A_{ap}}$ & RGB-D & \textbf{55.2} & 59.8 & 0.167 & \textbf{0.599} & 0.134 \\
    \end{tabular}}
    \vspace{1mm}
    \caption{Comparison of uni-modal and multi-modal generators. 
    }
\end{table}

\paragraphcond{Multi-Modality.} 
In \tabref{tab:semantic_consistency}, we compare our multi-modal approach (row 4) to the uni-modal variants where we estimate the depth maps (row 1) and the RGB images (row 2) by separately trained networks. We use Absolute Relative Error (AbsRel), Root Mean Square Error (RMSE), and Square Relative Error (SqRel) \cite{eigen2014depth} to measure the accuracy of the depth maps. Both the image quality and the accuracy of the depth maps are higher if our network is trained for both modalities. This shows that depth improves the image quality and that appearance improves the generated depth maps. 

\paragraphcond{Architectures of Generator.}
Furthermore, we compare our approach to an extension of the uni-modal method OASIS \cite{schonfeld_sushko_iclr2021} to RGB-D images (RGB-D OASIS). We use the generator from OASIS but modify it such that it generates RGB-D images. For a fair comparison, we use the discriminator and loss functions from our approach since an RGB-D discriminator does not perform well as we showed in \tabref{tab:arch}. Note that we do not use the adaptive perceptual loss for these experiments. Compared to our method, RGBD-OASIS (row 3 in \tabref{tab:semantic_consistency}) performs much worse both in terms of image quality and depth accuracy. A few qualitative results are shown in \figref{fig:examples}. 

Finally, we compare our generator with two modal-dependent decoders for appearance and geometry to a variant with a single 
decoder for appearance and geometry (4D Gen) in \tabref{tab:semantic_consistency} (row 5). While the semantic consistency (mIoU) is the same, the quality of the generated RGB and depth images is lower compared to our approach.

\section{Conclusion}

In this paper, we proposed an approach to increase the appearance diversity of an annotated dataset for RGB-D semantic segmentation. To this end, we addressed the new task of semantic RGB-D image synthesis and proposed a semantic consistent multi-modal generator, which comprises a modal-independent encoder and two modal-dependent decoders. It fully utilizes multi-modal information and produces semantic consistent and realistic RGB-D images. We have demonstrated that the proposed approach performs by a large margin better than uni-modal approaches and that it can be used to improve the accuracy of an RGB-D semantic segmentation method. While the experimental evaluation focused on semantic segmentation, we expect that the approach can also be used for other robotics-related tasks where the number of training images is limited.     

\section*{Acknowledgment}
The work was funded by the Deutsche Forschungsgemeinschaft (DFG, German Research Foundation) – GA 1927/9-1 (KI-FOR 5351).

{\small
\bibliographystyle{ieee_fullname}
\bibliography{egbib}
}

\end{document}